\documentclass[runningheads]{llncs}
\usepackage{graphicx}
\usepackage{multirow}
\usepackage{graphicx}
\usepackage{xcolor}

\usepackage{subcaption}
\usepackage{tikz}
\usepackage{comment}
\usepackage{amsmath,amssymb} 
\usepackage{color}
\usepackage{ulem}
\usepackage{hyperref}
\usepackage[accsupp]{axessibility}  

\usepackage[width=122mm,left=12mm,paperwidth=146mm,height=193mm,top=12mm,paperheight=217mm]{geometry}

\begin{document}
\pagestyle{headings}
\mainmatter

\title{Are all combinations equal? Combining textual and visual features with multiple space learning for text-based video retrieval} 

\titlerunning{Combining textual and visual features with multiple space learning}
\author{Damianos Galanopoulos \and
Vasileios Mezaris}

\authorrunning{D. Galanopoulos and V. Mezaris}

\institute{CERTH-ITI \\Thermi-Thessaloniki, Greece\\ \email{\{dgalanop,bmezaris\}@iti.gr}
}
\maketitle

\begin{abstract}
In this paper we tackle the cross-modal video retrieval problem and, more specifically, we focus on text-to-video retrieval. We investigate how to optimally combine multiple diverse textual and visual features into feature pairs that lead to generating multiple joint feature spaces, which encode text-video pairs into comparable representations. To learn these representations our proposed network architecture is trained by following a multiple space learning procedure.
Moreover, at the retrieval stage, we introduce additional softmax operations for revising the inferred query-video similarities.
Extensive experiments in several setups based on three large-scale datasets (IACC.3, V3C1, and MSR-VTT) lead to conclusions on how to best combine text-visual features and document the performance of the proposed network.\footnote{Source code is made publicly available at: \url{https://github.com/bmezaris/TextToVideoRetrieval-TtimesV}}

\end{abstract}

\section{Introduction}

Cross-modal information retrieval refers to the task where queries from one or more modalities (e.g., text, audio etc.) are used to retrieve items from a different modality (e.g., images or videos). This paper focuses on text-video retrieval, a key sub-task of cross-modal retrieval. The text-video retrieval task aims to retrieve unlabeled videos using only textual descriptions as input. This supports real-life scenarios, such as a human user searching for a video he/she remembers having viewed in the past, e.g., ``\textit{I remember a video where a dog and a cat were laying down in front of a fireplace}'', or, searching for a video never seen before, again by expressing their information needs in natural language, e.g., ``\textit{I would like to find a video where some kids are playing basketball in an open field'}'. 

To perform text-video retrieval, typically the videos or video parts, along with the textual queries, need to be embedded into a joint latent feature space. Early approaches to this task \cite{markatopoulou2017query} \cite{Video2vec} tried to annotate both modalities with a set of pre-defined visual concepts, and retrieval was performed by comparing these annotations. With the rise of deep neural networks over the past years, the community turned to them. Although various DNN architectures have been proposed to this end, their general strategy is the same: encode text and video into one or more joint latent feature spaces where text-video similarities can be calculated.

\begin{figure}
\centering
\includegraphics[width=.75\textwidth]{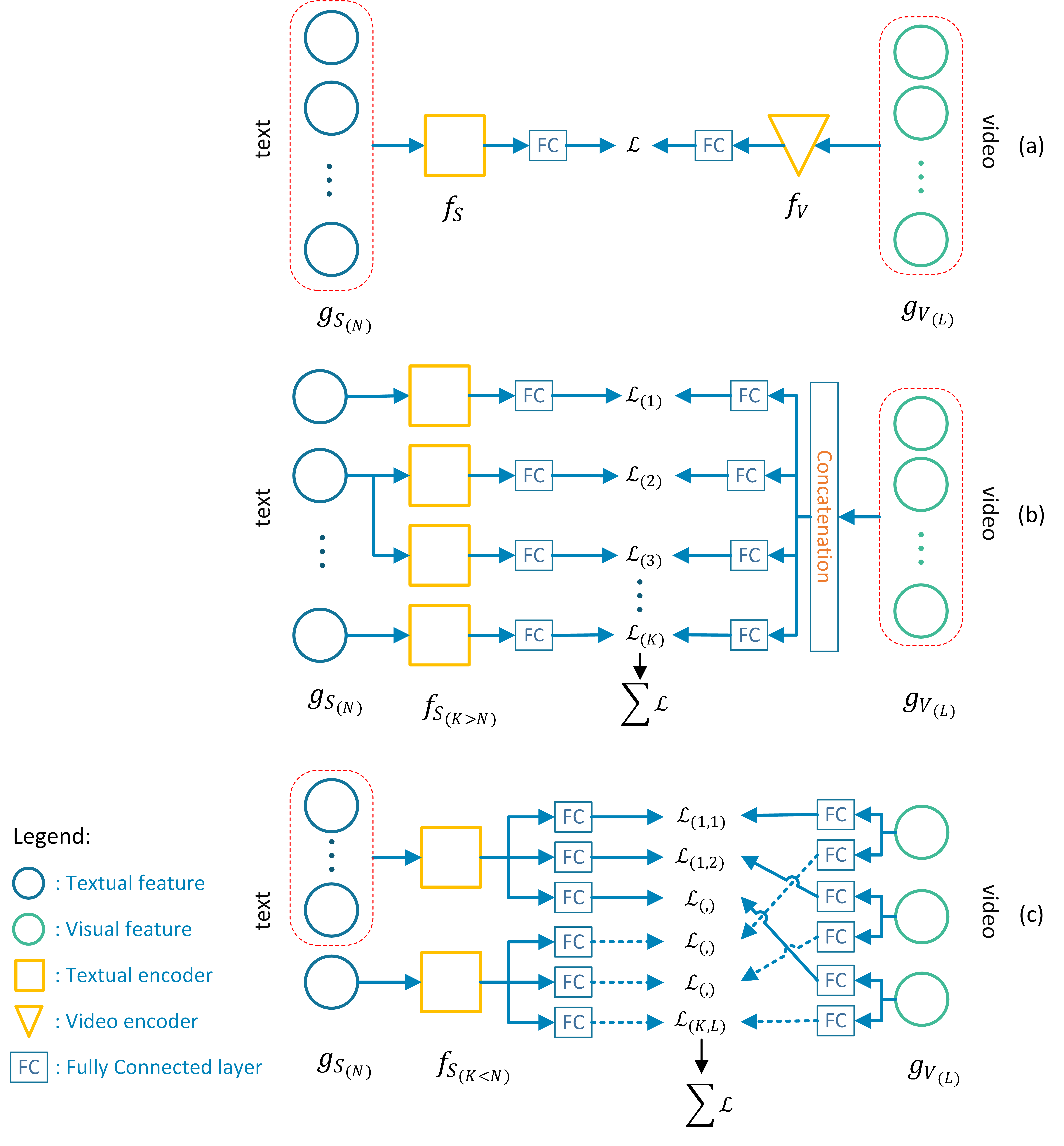}
\caption{Illustration of training different network architectures for dealing with the text-video retrieval problem using video-caption pairs. In all illustrations, $\mathcal{L}$ stands for the loss function. (a) All features are fed into one encoder per modality, (b) Every textual feature is used as input to a different encoder (or to more than one encoders), while visual features are simply concatenated, and (c) the proposed T$\times$V approach, where various textual and visual features are selectively combined to create different joint spaces}
\label{fig_1}
\end{figure}

State-of-the-art cross-modal video retrieval approaches utilize textual information by exploiting several textual features $g_s(\cdot)$ -- extracted with the help of already-trained deep networks or non-trainable extractors -- and encoding them through one or more trainable textual encoders (which are trained end-to-end as part of the overall cross-modal network training). The simple but widely used Bag-of-Words (bow) feature is often combined with embedding-based features such as Word2vec \cite{mikolov2013efficient} and Bert \cite{devlin2018bert}. Typically, these features are used as input to a simple or more sophisticated trainable textual encoder $f_s(\cdot)$, e.g, \cite{8353472}, \cite{li2019w2vv++}, that encodes them into one single representation. Similarly to textual information processing, trained image or video networks (e.g. a ResNet-152 trained on ImageNet) are used to extract feature vectors $g_v(\cdot)$ from the video frames. Typically these features are then concatenated and used as input to a trainable video encoder $f_v(\cdot)$. Finally, the outputs of $f_s(\cdot)$ and $f_v(\cdot)$ (after a linear projection and a non-linear transformation) are embedded into a new joint space (Fig. \ref{fig_1}a). Methods that follow this general methodology include \cite{dong2019dual}\cite{galanopoulos2020}\cite{li2019w2vv++}.

In \cite{li2020sea} a new approach of textual encoder assembly was proposed for exploiting diverse textual features. Instead of inputting all these features into a single textual encoder, an architecture where each textual feature is input into a different encoder (or to more than one encoders) was proposed, resulting in multiple joint latent spaces being created. However, when it comes to the video content, its treatment in \cite{li2020sea} is much simpler: several video features derived from trained networks are combined via vector concatenation, and individual fully connected layers embed them into a number of joint feature spaces. The cross-modal similarity, which serves as the loss function, is calculated by summing the individual similarities in each latent space. Fig. \ref{fig_1}b illustrates the above architecture.

In terms of loss function, the majority of the proposed works, e.g.\cite{dong2019dual}\cite{li2019w2vv++}\cite{galanopoulos2021}, utilize the improved marginal ranking loss introduced by \cite{faghri2018vse++}. This loss utilizes the hard-negative samples within a training batch to separate the positive samples from the samples that are negative but are located near to the positives. In \cite{cheng2021improving} the  dual softmax loss, a modification of the symmetric cross-entropy loss, was introduced. It is based on the assumption that the optimal text-video similarity is reached when the diagonal of a constructed similarity matrix contains the highest scores. So, this loss takes into consideration the cross-direction similarities within a training batch and revises the predicted text-video similarities.

In this work, inspired from \cite{li2020sea} where multiple textual encoders are introduced, we propose a new cross-modal network architecture to explore the combination of multiple and heterogeneous textual and visual features. We expand the textual information processing strategy of \cite{li2020sea}, with adaptations, to the visual information processing as well, and we propose a multiple latent space learning approach, as illustrated in Fig. \ref{fig_1}c. Moreover, inspired by the dual softmax loss of \cite{cheng2021improving}, we examine our network's performance when we introduce a dual softmax operation at the evaluation stage (contrarily to \cite{cheng2021improving} that applies it to the network's training) and use it to revise the inferred text-video similarity scores.

The contributions of this work are the following:
\begin{itemize}
\item We propose a new network architecture, named T$\times$V, to efficiently combine textual and visual features using multiple loss learning for the text-based video retrieval task.
\item We propose introducing a dual softmax operation at the retrieval stage for exploiting prior text-video similarities to revise the ones computed by the network.
\end{itemize}

\section{Related Work}

The general idea behind text-video retrieval is to project text and video information into comparable representations. Due to computational resources limitations, early approaches e.g. \cite{10.1145/2911996.2912015}\cite{markatopoulou2017query}\cite{habibianVideoStory2014}, dealt with relatively small datasets, and used pre-defined visual concepts as a stepping stone. I.e., videos and text were annotated with concepts, and text-video similarity was calculated by measuring the similarity between these annotations.
With the explosion of deep learning, the state-of-the-art moved forward to proposing concept-free methods.
The current dominant strategy is to encode both modalities into a joint latent feature space, where the text and video representations are compared. 

In \cite{dong2021dual} and \cite{galanopoulos2020}, dual encoding networks were proposed. Two similar sub-networks were introduced, one for the video stream and one for the text, to encode them into a joint feature space. In the dual-task network of \cite{wu2020}, a combination of latent space encoding and concept representation, was proposed: the first task encodes text and video into a joint latent space, while the second task encodes video and text as a set of visual concepts. In \cite{li2019w2vv++} several textual features were used to create multiple textual encoders, instead of feeding them into a single encoder. In this way, multiple joint  text-video latent feature spaces could be learned, leading to more accurate retrieval results. In \cite{song2019polysemous} the problem of understanding textual or visual content with multiple meanings is addressed by combining global and local features through multi-head attention. More recently, inspired by the human reading strategy, \cite{dong2022reading}  proposed a two-branches approach to encode video representations. A preview branch captures the overview information of a video, while the intense-reading branch is designed to extract more in-depth information. Moreover, the two branches interact, and the preview guides the intense-reading branch. As a general trend, the various recent works on text-video retrieval, e.g. \cite{li2020sea}\cite{dong2022reading}\cite{chen2021matters}, have shown that the utilization of multiple textual features to create more than one video-text joint spaces leads to improved overall performance.

Recent approaches,  additionally go beyond the standard evaluation protocol (i.e., training the network using the training portion of a dataset and testing it on the testing portion), benefiting from pre-training on further large-scale video-text datasets. This procedure leads to improved performance and learning transferable textual and visual representations. In \cite{miech2019howto100m}, HowTo100M is introduced: a large-scale dataset of ~100M web videos. Using this dataset to pre-train a baseline video retrieval network is shown in \cite{miech2019howto100m} to be beneficial. HiT \cite{liu2021hit} uses a transformer-based architecture to create a hierarchical cross-modal network for creating semantic-level and feature-level encoders. In \cite{liu2021hit} experimentation with and without pre-training also shows that the network's performance increases with the pre-training step. BridgeFormer \cite{ge2022bridging} introduces a module that is trained to answer textual questions in order to be used as the pre-training step of a dual encoding network. Frozen \cite{bain2021frozen}, on the other hand, is based on a transformer architecture and does not use trained image DNNs as feature extractors. It did introduce, though, a large-scale video-text dataset (WebVid-2M) which was used for end-to-end pre-training of their network.

\section{Proposed Approach}

\subsection{Overall Architecture}

The text-video retrieval problem is formulated as follows: let $V = \{v_1,v_2,\ldots,v_T\}$ be a large set of $T$ unlabeled video shots and $s$ a free-text query. The goal of the task is, given the query $s$, to retrieve from $V$ a ranked list with the most relevant video shots.

Our T$\times$V network consists of two key sub-networks, one for the textual and one for the visual stream. The textual sub-network inputs a free-text query and vectorizes it into $M$ textual features $g_S: \{g_{s}^{1}(\cdot), g_{s}^{2}(\cdot), \dotsc,g_{s}^{M}(\cdot)\}$. These $M$ features are used as input in a set of carefully-selected $K$ textual encoders $f_S: \{f_{s}^{1}(\cdot), f_{s}^{2}(\cdot), \dotsc,f_{s}^{K}(\cdot)\}$ that encode the input sentence. Each of these encoders can be either a trainable network or simply an identity function that just forwards its input.
Similarly to the textual one, the visual sub-network inputs a video shot consisting of a sequence of $N$ keyframes ${v} = \{I_1,I_2,\ldots,I_N\}$. We use $L$  trained DNNs in order to extract the initial frame representations $g_V: \{g_{v}^{1}(\cdot), g_{v}^{2}(\cdot), \dotsc,g_{v}^{L}(\cdot)\}$. To obtain video-shot level representations we follow the mean-pooling strategy.

Subsequently, we create all the possible textual encodings-visual feature pairs ($f_{s}^{k}(s)$, $g_{v}^{l}(v)$) and a joint embedding space is created for each pair, using to this end two fully connected layers. Thus, $ K\times L$ different joint spaces are created. The objective of our network is to learn a similarity function $similarity(s,v)$ that will consider every individual similarity in each joint latent space utilizing multi-loss-based training. Fig. \ref{fig_1}c illustrates our proposed method.

\subsection{Multiple Space Learning}

To encode the $(f_{s}^{k}(\cdot),g_{v}^{l}(\cdot))$ pair into its joint feature space, as shown in Fig. \ref{fig_1}c,  each single part of the pair is linearly transformed by a fully connected layer (FC). A non-linearity is added in the FC output (not illustrated in Fig. \ref{fig_1}c for brevity), for which the ReLU activation function is used, as follows:

\begin{align*} \mathbf{s}_k = ReLU(FC(f_{s}^{k}(\cdot)) \\\mathbf{v}_l = ReLU(FC(g_{v}^{l}(\cdot)) \end{align*}

This transformation encodes the $(f_{s}^{k}(\cdot),g_{v}^{l}(\cdot))$ pair into its new joint feature space. The similarity function $sim(\mathbf{s}_k,\mathbf{v}_l)$ calculates the similarity between the output of textual encoder $k$ and video feature $l$ in this joint feature space. The overall similarity between a video-sentence pair is calculated as follows:

\begin{align*}
similarity(s,v) = \sum_{k=1}^{K}\sum_{l=1}^{L} sim({\mathbf{s}_k,\mathbf{v}_l})\end{align*}

where  $sim(\mathbf{s}_k,\mathbf{v}_l) = cosine\_similarity(\mathbf{s}_k,\mathbf{v}_l)$.

To train our network, similarly to \cite{galanopoulos2020}\cite{dong2021dual}, we utilize the improved marginal ranking loss introduced in \cite{faghri2018vse++}. This emphasizes on the hard-negative samples in order to learn to maximize the similarity between textual and video embeddings. At the training stage, given a sentence-video sample $(s,v)$, for a specific latent feature space $(k,l)$, the improved marginal loss is defined as follows:

\begin{align*}
\mathcal{L}_{(k,l)}(s,v) = max(0, \alpha + sim(\mathbf{s}_{k},\mathbf{v}_{l}^{'}) - sim(\mathbf{s}_k, \mathbf{v}_{l}))\\ + max(0,  \alpha + sim(\mathbf{s}_{k}^{'}, \mathbf{v}_{l}) - sim(\mathbf{s}_k,\mathbf{v}_{l}))   
\end{align*}

where $\mathbf{v}_{l}^{'}$ and $\mathbf{s}_{k}^{'}$ are the hardest negatives of $\mathbf{s}_{l}$ and $\mathbf{v}_{k}$ respectively and $ \alpha$  is a hyperparameter for margin regulation. The overall training loss is calculated as the sum of all $K \times L$ individual loss values:

\begin{align*}
\mathcal{L}(s,v) = \sum_{k=1}^{K}\sum_{l=1}^{L}\mathcal{L}_{(k,l)}(s,v) 
\end{align*}

\subsection{Dual Softmax Inference}

\begin{figure}
\centering

\includegraphics[width=\textwidth]{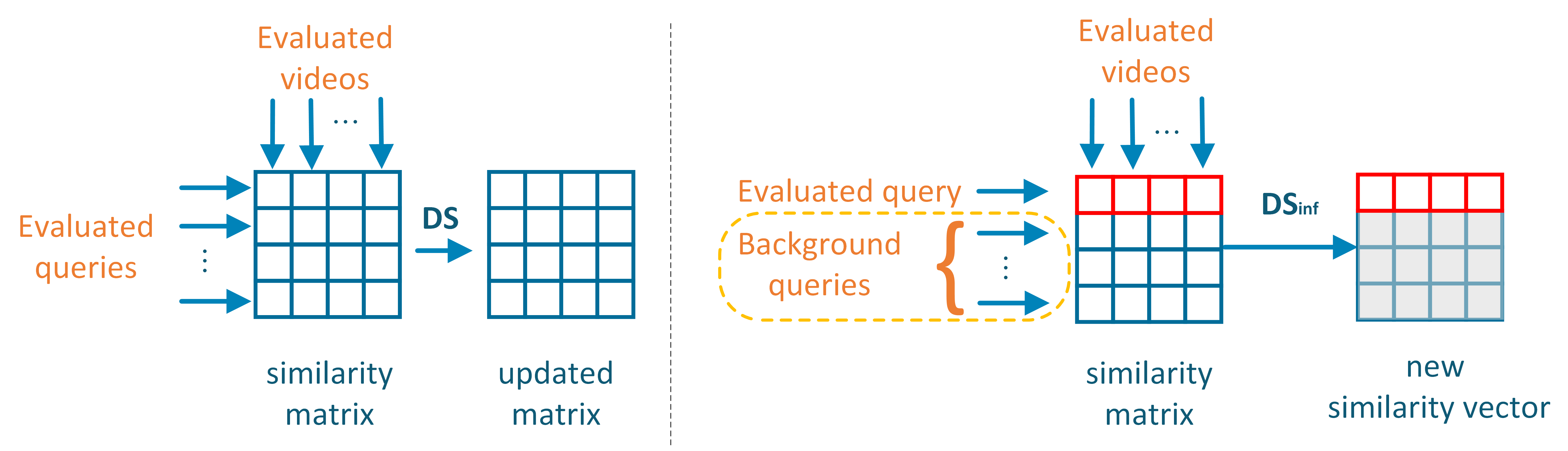}
\caption{Different approaches to update the similarity matrix using all the evaluated queries (left subfigure) and pre-defined background queries (right subfigure)}
\label{fig:fig_3a}
\end{figure}

In \cite{cheng2021improving}, a new objective function based on two softmax operations was proposed. According to this, at the training stage the predicted text-video similarities were revised by calculating a so-called \textit{cross-direction} similarity matrix and multiplying it with the predicted one. Specifically, during a training batch, let $Q$ be the number of examined caption-video pairs. By computing the similarities between every caption and all videos, a similarity matrix $\mathbf{X} \in \mathcal{R}^{Q\times Q}$ was generated. Next, by applying two cross-dimension softmax operations (one column-wise and one row-wise) an updated similarity matrix $\mathbf{X}'$ was calculated, and was subsequently used as discussed in \cite{cheng2021improving}. Directly applying this approach at the inference stage, though, would require that all queries to be evaluated are known a priori and are evaluated simultaneously; they would need to be used for calculating matrix $\mathbf{X}$, as illustrated in the left part of Fig. \ref{fig:fig_3a}. This is not a realistic scenario, especially in real-world retrieval applications. 

\begin{figure}
\centering

\includegraphics[width=0.85\textwidth]{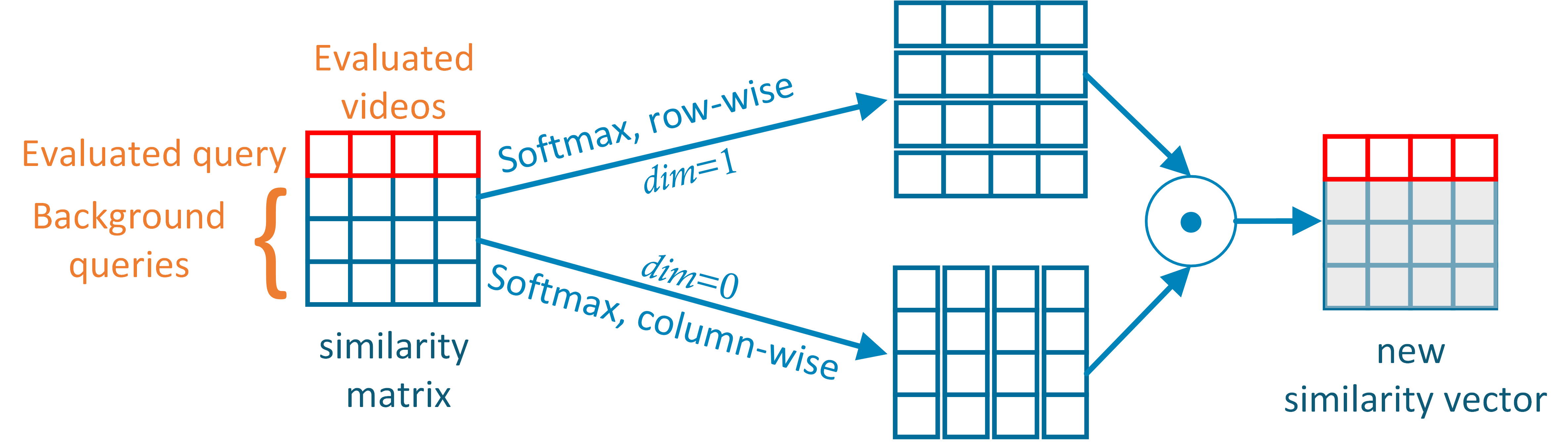}
\caption{Illustration of the dual softmax-based inference (DS\textsubscript{inf}) approach for updating the similarities }
\label{fig_4}
\end{figure}

To deal with this issue and revise the inferred text-video similarities at the retrieval stage, we propose a dual softmax-based inference (DS\textsubscript{inf}) as illustrated in the right side of Fig. \ref{fig:fig_3a}. We utilize a fixed set of  $C$ pre-defined background textual queries, which are independent of the evaluated dataset, and we calculate their similarities with all $D$  videos of the test set. For the set of background queries, we calculate once the similarity matrix $\mathbf{X}^{ \ast}\in \mathcal{R}^{C\times D}$. For each individual evaluated query $s$ a similarity vector $\mathbf{y}(s) = [similarity(s,v_1), similarity(s,v_2), \allowbreak \ldots, similarity(s,v_D)]^T$, is calculated. A matrix $\mathbf{Z}(s) = concat(\mathbf{y}(s);\mathbf{X}^{ \ast})$ is constructed, and a dual softmax operation revises the similarities as follows:

\begin{align*}
    \mathbf{Z}^{*}(s) = Softmax(\mathbf{Z}(s),\ dim=0) \odot Softmax(\mathbf{Z}(s), \ dim=1) 
\end{align*}
where $ \odot $ denotes the Hadamard product. Finally, from matrix $\mathbf{Z}^{*}$ we extract the revised similarity vector $\mathbf{y}^{*} = [{Z}^{*}_{0,1}, {Z}^{*}_{0,2},\cdot,{Z}^{*}_{0,D}]$ (Fig. \ref{fig_4}). This normalization procedure is meaningful when we expect that there are multiple positive video samples in our dataset for the evaluated query; thus, by normalizing the inferred similarities we can produce a better ranking list.

\subsection{Specifics of Textual Information Processing}

In this section, we present every textual feature and every textual encoder we used in order to find the optimal textual encoder combination. Given a sentence $s$ consisting of $\{w_1, w_2, \ldots, w_B\}$ words, we utilize $M=4$ different textual features. These features are used as input to textual encoders.

\paragraph{\textbf{Textual features}}
\begin{itemize}
\item Bag-of-Words (bow): We utilize Bag-of-Words to vectorize every sentence into a sparse vector representation expressing the occurrence frequency of every word from a pre-defined vocabulary.

\item Word2Vec (w2v): Word2Vec model \cite{mikolov2013efficient} is an established and well-performing word embedding model. W2v learns to embed words into a word-level representation vectors $\mathbf{W}_{w2v}:\{\mathbf{w}_1^{w2v}, \mathbf{w}_2^{w2v}, \ldots, \mathbf{w}_B^{w2v} \}\in R^{BxD_{w2v}}$. The overall sentence w2v embedding $g_{s}^{w2v} $ is calculated as the mean pooling of the individual word embeddings.

\item Bert:  Bert \cite{devlin2018bert} offers contextual embeddings by considering the sequence of all words inside a sentence, which means that the same word may have different embedding when the context of a sentence is different. We utilize the BASE variation of bert consisting of 12 encoders and 12 bidirectional self-attention heads. We calculate the bert sentence embedding $g_{s}^{bert} $ by mean pooling the individual word embedding.

\item Clip: The transformer-based trained model of CLIP \cite{clip_2021} is used as a textual feature extractor. Sentence $s$ is fed to it as a sequence of words $w_1,..., w_B$ and token embeddings $\mathbf{W}_{clip}:\{ \mathbf{w}_{startoftext}^{clip}, \mathbf{w}_1^{clip}, \ldots, \mathbf{w}_B^{clip}, \mathbf{w}_{endoftext}^{clip} \}$ are calculated. The last token embedding, $\mathbf{w}_{endoftext}^{clip}\in R^{512}$, is used as our feature vector. 

 \end{itemize}

\paragraph{\textbf{Textual Encoders}}
\begin{figure}
\centering
\includegraphics[width=\textwidth]{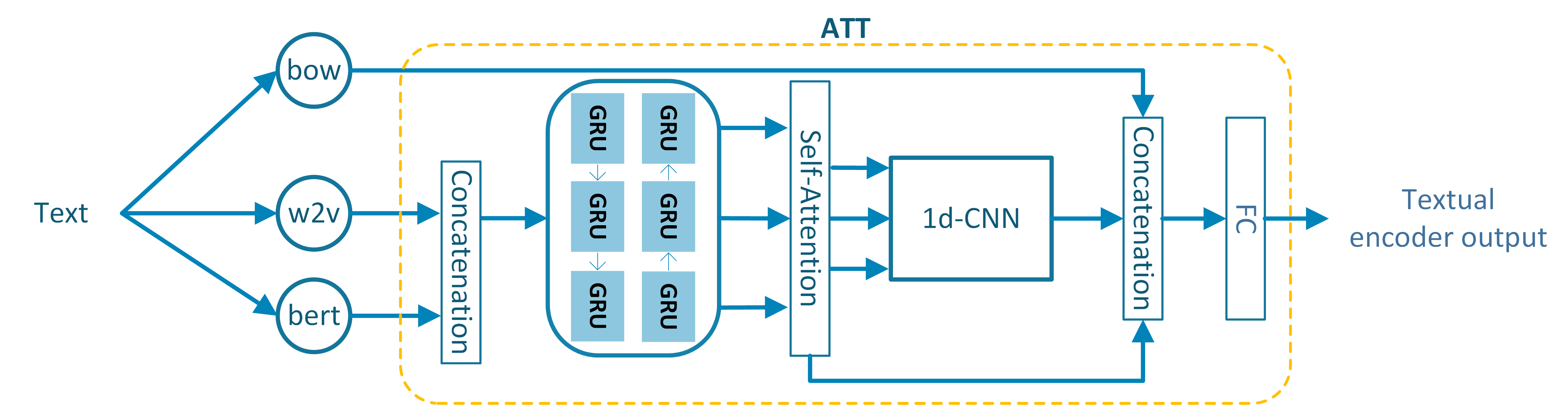}
\caption{ Illustration of the ATT encoder, inputting several textual features and producing three levels of encodings (bow, self-attention bi-gru and CNN outputs) which are concatenated and contribute to the final output of the encoder}
\label{fig_2}
\end{figure}

The textual encoders input the extracted textual features (one at a time, or a combination of them) and output a new embedding. We experimented with combinations of the followings:

\begin{itemize}
\item $f_{s}^{bow}$, $f_{s}^{w2v}$, $f_{s}^{bert}$, $f_{s}^{clip}$:  these encoders feedforward the corresponding features through an identity layer.
\item $f_{s}^{w2v-bert}$: The concatenation of w2v and bert features is used to feed an identity layer.
\item $f_{s}^{bi-gru}$: A self-attention bi-gru module, introduced in \cite{galanopoulos2020}, is trained as part of the complete network architecture; it takes as input the w2v features for each word and their temporal order (i.e., not using the overall sentence w2v embedding, contrarily to $f_{s}^{w2v}$ and $f_{s}^{w2v-bert}$).
\item Attention-based dual encoding network (ATT):
The textual sub-network presented in \cite{galanopoulos2020} (illustrated in Fig. \ref{fig_2}) is trained (similarly to $f_{s}^{bi-gru}$, above), taking as input the bow, w2v and bert features (again, for each individual word rather than the mean-pooled sentence embeddings) and producing a vector in $\mathcal{R}^{2048}$.

\end{itemize}

Among all the above possible textual encoders, we propose to combine in our network the $f_{s}^{clip}$ and ATT ones, as experimentally verified in Section \ref{sec:exps}.

\subsection{Specifics of Visual Information Processing}

Similarly to the textual sub-network, we use several deep networks that have been trained for other visual tasks as frame feature extractors. Considering a video shot $v$, first we uniformly sample it with a fixed rate of 2 frames per second, resulting in a set of keyframes $\{I_1,I_2,\ldots,I_N\}$. Then, frame-level representations are obtained with the help of the feature extractors listed below, followed by mean pooling of the individual frame representations to get shot-level features. 

\paragraph{\textbf{Visual features}}
\begin{itemize}
\item R\_152: The first video feature extractor inputs an image frame into a ResNet-152 \cite{he2016deep} network, trained on the ImageNet-11k dataset. The flattened output of the layer before the last fully connected layer of the network is used as a feature representation of every frame in $\mathcal{R}^{2048}$. 

\item Rx\_101: The second feature extractor utilizes a ResNeXt-101 network, pre-trained by weakly supervised learning on web images followed and fine-tuned on ImageNet \cite{mahajan2018exploring}. Similarly to R\_152, Rx\_101 inputs frames and the frame representations in $\mathcal{R}^{2048}$, are obtained as the flattened output of the layer before the last fully connected layer. 
\item Clip: As third video feature extractor we utilise a trained CLIP model (ViT-B/32) \cite{clip_2021}, to create frame-level representations in $\mathcal{R}^{512}$. 

\end{itemize}

As illustrated in Fig. \ref{fig_1}c, we propose using these visual features, without introducing any trainable visual encoder, directly as input to a number of individual FC layers for learning the latent feature spaces. However, in order to examine more design choices, we also tested in our ablation experiments the introduction of a visual encoder, similarly to what we do for textual information processing. To this end, we utilized the visual sub-network of the attention-based dual encoding network of \cite{galanopoulos2020} (ATV). Following \cite{galanopoulos2020}, we input all three aforementioned frame-level features to a single ATV encoder, which (similarly to the ATT one) was trained end-to-end as part of our overall network.


\section{Experimental Results} \label{sec:exps}

\subsection{Datasets and Experimental Setup}

\begin{table}[]
\centering
\caption{Results and comparisons on the IACC.3 and V3C1 datasets. Bold/underline indicates the best-/second-best scores}
\label{tab:tabavs_3}
\resizebox{0.85\columnwidth}{!}{%
\begin{tabular}{lccccccc|c}
\hline
\multicolumn{1}{c}{\multirow{3}{*}{Model}} & \multicolumn{1}{c}{} & \multicolumn{6}{c}{Datasets (all scores: MxinfAP)} &  \\ \cline{3-8} 
\multicolumn{1}{c}{} & \multicolumn{1}{c}{}  & \multicolumn{3}{c|}{IACC.3} & \multicolumn{3}{c}{V3C1} &  \\ \cline{3-8} 
\multicolumn{1}{c}{} & \multicolumn{1}{c}{Year}  &\multicolumn{1}{c}{AVS16} & \multicolumn{1}{c}{AVS17} & \multicolumn{1}{c|}{AVS18} & \multicolumn{1}{c}{AVS19} & \multicolumn{1}{c}{AVS20} & \multicolumn{1}{c}{AVS21}  & Mean \\ \hline

Dual Encoding \cite{dong2019dual} & 2019 & 0.165   & 0.228    &  \multicolumn{1}{c|}{0.117} &  0.152 &  -  & - & - \\

ATT-ATV \cite{galanopoulos2020} & 2020 & 0.164 & 0.243 & \multicolumn{1}{c|}{0.127} & - & - & - & - \\
ATT-ATV \cite{galanopoulos2020} (re-training)  & - & 0.202 & 0.281 & \multicolumn{1}{c|}{0.146} & \underline{0.208} & 0.283 & \underline{0.289} & 0.235\\ 
Dual-task \cite{wu2020} &  2020 & 0.185 & 0.241 & \multicolumn{1}{c|}{0.123} & 0.185 & - & - & - \\
SEA \cite{li2020sea} & 2021 & 0.164  & 0.228  & \multicolumn{1}{c|}{0.125} & 0.167 & - & - & - \\
SEA \cite{li2020sea} (re-training) & - & \underline{0.207} & 0.279 & \multicolumn{1}{c|}{0.148} & 0.191 & 0.283 & 0.283 & 0.232 \\

SEA-clip \cite{chen2021matters} & 2021 & 0.203 & \textbf{0.321} & \multicolumn{1}{c|}{\textbf{0.156}} & 0.192 & \textbf{0.329} & - & - \\
Extended Dual Encoding  \cite{dong2021dual} & 2022 & 0.159  & 0.244  &   \multicolumn{1}{c|}{0.126} & - & - & - & - \\  
 \hline
 T$\times$V (proposed)&  & \textbf{0.234} & \underline{0.317} & \multicolumn{1}{c|}{\underline{0.153}} & \textbf{0.220 }& \underline{0.316} &\textbf{ 0.312} & \textbf{0.259} \\

\hline
\end{tabular}%
}
\end{table}

We evaluate our approach and report experimental results on three datasets: the two official TRECVID AVS datasets (i.e., IACC.3 and V3C1) \cite{2021trecvidawad} and MSR-VTT \cite{xu2016msr}. The AVS datasets are designed explicitly for text-based video retrieval evaluation; they include the definition of tens of textual queries as well as ground-truth associations of multiple positive samples with each query. The IACC.3 dataset consists of $335.944$ test videos, and the V3C1 of $1.082.629$ videos for testing (most of which are not associated with any of the textual queries). As evaluation measure we use the mean extended inferred average precision (MxinfAP), as proposed in \cite{2021trecvidawad} and typically done when working with these datasets. On the other hand, MSR-VTT targets primarily video captioning evaluation, but is also often used for evaluating text-video retrieval methods. It is made of $10.000$ videos and each video is annotated with $20$ different natural language captions (totaling $200.000$ captions, which are generally considered to be unique); for retrieval evaluation experiments, given a caption the goal is to retrieve the single video that is ground-truth-annotated with it. For the MSR-VTT experiments, following the relevant literature, we use as evaluation measures the recall $R@k$, $k=1,5,10$, the median rank (\textit{Medr}) and mean average precision (\textit{mAP}).

Regarding the training/testing splits: for the evaluations on the AVS datasets, our cross-modal network (and any network of the literature, i.e \cite{galanopoulos2020}, \cite{li2020sea}, that we re-train for comparison) is trained using a combination of four other large-scale video captioning datasets: MSR-VTTT \cite{xu2016msr}, TGIF \cite{li2016tgif}, ActivityNet \cite{caba2015activitynet} and Vatex \cite{wang2019vatex}. For validation purposes, during training, we use the Video-to-Text Description dataset of TRECVID 2016. For testing, all sets of queries specified by NIST for IACC.3 (i.e., AVS16, AVS17 and AVS18) and V3C1 (i.e., AVS19, AVS20 and AVS21) are used. For the evaluations on the MSR-VTT dataset, we experimented with two versions of this dataset: MST-VTT-full \cite{xu2016msr} and MSR-VTT-1k-A \cite{yu2018joint}. MST-VTT-full consists of $6.513$ videos for training, $497$ for validation and $2.990$ videos (thus, $2.990\times 20$ video-caption pairs) for testing. On the other hand, MSR-VTT-1k-A contains $9.000$ videos for training and $1.000$ video-caption pairs for testing. For both MSR-VTT versions, we trained our network of the training portion of the dataset and report results on the testing portion, respectively.

Regarding the training conditions and parameters: To train the proposed network, (and again, also for re-training \cite{galanopoulos2020}, \cite{li2020sea}) we adopt the setup of \cite{galanopoulos2021}, where six configurations of the same architecture with different training parameters were combined. Specifically, each model is trained using two optimizers, i.e., Adam and RMSprop, and three learning rates ($1\times10^4$, $5\times10^5$, $1\times10^5$). The final results for a given architecture are obtained by combining the six returned ranking lists of the individual configurations in a late fusion scheme, i.e. by averaging the six obtained ranks for each video. For training all configurations, we follow a learning rate decay technique, and we reduce the learning rate $1\%$ per epoch or by $50\%$ if the validation performance does not improve for three epochs. The dropout rate is set to 0.2 to reduce overfitting. Also, following \cite{faghri2018vse++} the margin parameter on loss function is set to $\alpha=0.2$. All experiments were performed on a single computer equipped with Nvidia RTX3090 GPU. Our models were implemented and trained using Pytorch 1.11.

\begin{table}

\centering
\caption{Results and comparisons between different encoder strategies when using dual softmax-based inference (DS\textsubscript{inf}) similarity. Scores in bold/underline indicate the best-/second-best-performing strategy}
\label{tab:tabavs_2}
\resizebox{0.85\columnwidth}{!}{%
\begin{tabular}{l|lccc|ccc|c} 
\hline
\multicolumn{2}{c}{\multirow{3}{*}{Model}} & \multicolumn{6}{c}{Datasets (all scores: MxinfAP)} &  \\ 
\cline{3-8}
\multicolumn{2}{c}{} & \multicolumn{3}{c|}{IACC.3} & \multicolumn{3}{c}{V3C1} &  \\ 
\cline{3-9}
\multicolumn{2}{c}{} & AVS16 & AVS17 & AVS18 & AVS19 & AVS20 & AVS21 & Mean \\ 
\hline
\multicolumn{2}{l}{ T$\times$V} & 0.234 & 0.317 & 0.153 & 0.220 & 0.316 & 0.312 & 0.259 \\
\multicolumn{2}{l}{\begin{tabular}[c]{@{}l@{}} T$\times$V + DS\textsubscript{inf} on the \\ set of evaluated queries\end{tabular}} & \textbf{0.244} & \textbf{0.330} & \textbf{0.165} & \textbf{0.226} & \underline{0.324} & \textbf{0.324} & \textbf{0.269} \\ 
\cline{1-2}
\multirow{4}{*}{\begin{tabular}[c]{@{}l@{}} T$\times$V + DS\textsubscript{inf} \\ using as \\ background \\queries:\end{tabular}} & 60 random captions & 0.240 & 0.323 & 0.157 & 0.223 & 0.318 & 0.313 & 0.262 \\
 & 200 random captions & 0.239 & 0.322 & 0.158 & 0.224 & 0.315 & 0.315 & 0.262 \\
 &\begin{tabular}[c]{@{}l@{}}other years' queries\\ of the same dataset\end{tabular}  & \underline{0.243} & \underline{0.328} & \underline{0.162} & \underline{0.226} & \textbf{0.325} & \underline{0.323} & \underline{0.268} \\
\hline
\end{tabular}
}
\end{table}

\subsection{Results and Comparisons}

Table \ref{tab:tabavs_3} presents the results of the proposed T$\times$V network, i.e. using three visual features (R\_152,  Rx\_101, clip) followed by FC layers and clip and ATT as textual encoders, on IACC.3 and V3C1 datasets and comparisons with state-of-the-art literature approaches. We compare our method with six methods and the presented results are extracted from their original papers. Furthermore, as \cite{chen2021matters} has shown that the quality of the initial visual features is crucial for the performance of a method, we present results of re-training the ATT-ATV \cite{galanopoulos2020} and SEA \cite{li2020sea} networks using the same visual features and same training datasets as we did in our experiments, using their publicly available code. Our proposed network outperforms the competitors on AVS16, AVS19 and AVS21. SEA-clip \cite{chen2021matters} achieves better results on AVS17, AVS18 and AVS20 by exploiting 3D CNN-based visual features. Comparing the mean performance on AVS16-AVS20 (since SEA-clip does not report results on AVS21), our network achieves MxinfAP equal to $0.248$ while SEA-clip $0.240$.

\begin{table}[]
\caption{Results and comparisons on the MSR-VTT full and 1k-A datasets. Methods marked with $^\ast$ use an alternative training set of 7.010 video-caption samples for the 1k-A dataset, but still report results on the same test portion of 1k-A as all other methods. Bold/underline indicates the best-/second-best scores}
\label{tab:tabmsr_1}
\resizebox{\columnwidth}{!}{%
\begin{tabular}{llcccccccccc}
\hline
                                                   & & \multicolumn{10}{c}{Datasets}                                                                      \\ \cline{3-12} 
                                                   & & \multicolumn{5}{c}{MSR-VTT full}                                      & \multicolumn{5}{c}{MSR-VTT 1k-A}            \\ \cline{3-12} 
\multicolumn{1}{c}{Model}    & Year                   & R@1$\uparrow$  & R@5$\uparrow$  & R@10$\uparrow$ & Medr$\downarrow$ & \multicolumn{1}{l|}{mAP$\uparrow$} & R@1$\uparrow$ & R@5$\uparrow$ & R@10$\uparrow$ & Medr$\downarrow$ & mAP$\uparrow$   \\ \hline
W2VV  \cite{8353472}        &    2018           & 1.1  & 4.7  & 8.1  & 236  & \multicolumn{1}{c|}{3.7 } &   1.9  & 9.9    & 15.2     & 79     & 6.8\\
VSE++ \cite{faghri2018vse++}  & 2018     & 8.7  & 24.3 & 34.1 & 28   & \multicolumn{1}{c|}{16.9 } & 16.0    &    38.5 &  50.9    & 10     & 27.4      \\

CE \cite{liu2019use}    &      2019                 & 7.9  & 23.6 & 34.6 & 23   & \multicolumn{1}{c|}{16.5 } &  17.2   & 46.2    & 58.5     &    7  &    30.3       \\
W2VV++\cite{li2019w2vv++}   &    2019    & 11.1 & 29.6 & 40.5 & 18   & \multicolumn{1}{c|}{20.6 } &  19.0   &    45.0 &  58.7    &   7   & 31.8       \\

TCE \cite{yang2020tree}     &      2020              & 9.3  & 27.3 & 38.6 & 19   & \multicolumn{1}{c|}{18.7 } &  17.8   & 46.0    & 58.3     & 7     & 31.1         \\
HGR \cite{chen2020fine}      &       2020              & 11.1 & 30.5 & 42.1 & 16   & \multicolumn{1}{c|}{20.8} & 21.7    & 47.4    &    61.1  & 6     & 34.0        \\
SEA \cite{li2020sea} $^\ast$      &    2021     & 12.4 & 32.1 & 43.3 & 15   & \multicolumn{1}{c|}{22.3 } &   23.8  &    50.3  & 63.8     & 5      & 36.6       \\

HiT \cite{liu2021hit}  &  2021 & - & - &  - &   -  &   \multicolumn{1}{c|}{ -} & 27.7  &    59.2      &  72.0  &      3  &   -    \\
HiT (Pre-train. on HT100M)  \cite{liu2021hit}  &  2021  & - & - & -  & -   &    \multicolumn{1}{c|}{- } &  30.7   &  60.9     & 73.2    &   \underline{2.6}     &   -   \\

CLIP \cite{portillo2021straightforward}$^\ast$   &           2021                     &\textbf{21.4}  & \underline{41.1}   & \underline{50.4}  &  \underline{10}   &    \multicolumn{1}{c|}{ -} &   31.2   &   53.7  &   64.2    &    4   &   -    \\
FROZEN  \cite{bain2021frozen}        &    2021                       & - &  - & - &   -  &    \multicolumn{1}{c|}{- } &  31.0 &  59.5   &70.5  & 3       & -             \\
Extended Dual Encoding \cite{dong2021dual}$^\ast$  & 2022  & 11.6 & 30.3 & 41.3 & 17   & \multicolumn{1}{c|}{21.2} &  21.1   &  48.7   &    60.2  & 6     & 33.6        \\
RIVRL \cite{dong2022reading} $^\ast$      &     2022            & 13.0 & 33.4 & 44.8 & 14   & \multicolumn{1}{c|}{23.2} &  23.3   & 52.2    &    63.8  & 5     & 36.7        \\
RIVRL with bert \cite{dong2022reading} $^\ast$    &       2022                         & 13.8 & 35.0 & 46.5 & 13   & \multicolumn{1}{c|}{\underline{24.3}} &  27.9   & 59.3    & 71.2     & 4     & 42.0       \\
BridgeFormer  \cite{ge2022bridging}     &  2022                             & - &  - & - & -    &    \multicolumn{1}{c|}{ -} &  \textbf{37.6} &  \underline{64.8}   & 73.1  & 3       & -             \\
 \hline
  T$\times$V (proposed) $^\ast$   &     & -   &  - &  - &   -  &    \multicolumn{1}{c|}{- }  &   32.3   & 63.7    &    \underline{74.6}   &   3   & \underline{46.3}        \\ 
 T$\times$V (proposed)  &      & \underline{21.2}    &  \textbf{46.3}   &\textbf{58.2}    &   \textbf{7}   &   \multicolumn{1}{c|}{\textbf{33.1}}   & \underline{36.5}   &   \textbf{66.9} &  \textbf{77.7}   &   \textbf{2}   &  \textbf{50.2}  \\ \hline
\end{tabular}
}
\end{table}

In Table \ref{tab:tabavs_2} we experiment with the dual softmax-based (DS\textsubscript{inf}) inference on IACC.3 and V3C1. We examine the impact of different background query strategies and compare with the proposed network without DS\textsubscript{inf}. ``\textit{DS\textsubscript{inf} on the set of evaluated queries}'' indicates that the operation was performed using the same year's queries. For example, to retrieve videos for an AVS16 query, the remaining AVS16 queries are used as background queries. This improves the overall performance, but as we already have discussed, is not a realistic application scenario. To overcome this problem, we experiment with using a fixed set of background queries: 60 or 200 randomly selected captions extracted from the training datasets. By using these captions, performance is improved compared to the proposed network in every dataset and every year. The difference between using 60 and 200 captions is marginal. Finally, we try using as background queries all AVS queries defined for the same video dataset but not the same test-year as the examined query. For example, to evaluate each of the AVS16 queries, we use the AVS17 and AVS18 queries as background. This strategy achieves the second-best results among all examined ones, being marginally outperformed by ``\textit{DS\textsubscript{inf} on the set of evaluated queries}'' strategy; and, contrarily to the latter, does not assume knowledge of all the evaluation queries beforehand. 

In Table \ref{tab:tabmsr_1} we present results on the MSR-VTT full and 1k-A datasets and compare with literature methods. Our network outperforms most literature methods, even methods like FROZEN \cite{bain2021frozen} and HiT (pre-trained on HT100M) \cite{liu2021hit} in which their networks utilize a pre-training step on other large text-video datasets. Moreover, BridgeFormer \cite{ge2022bridging}, using a pre-training step on the WebVid-2M \cite{bain2021frozen} dataset, marginally outperforms our network in $R@1$ terms, while our approach achieves better results on the remaining evaluation measures. Finally, we should note that experiments with DS\textsubscript{inf} on MSR-VTT (not shown in Table \ref{tab:tabmsr_1}) lead to only marginal differences in relation to the results of the proposed T$\times$V network. This is expected because of the nature of MSR-VTT: given a caption the goal is to retrieve the {\it single} video that is ground-truth-annotated with it; thus, re-ordering the entire ranking list by introducing the DS\textsubscript{inf} normalization of the caption-video similarities has limited impact.

\subsection{Ablation Study}

In this section we study the effectiveness of different textual encoders and visual features (or also encoders) combinations. We report results using three visual feature encoding strategies: ``\textit{feat. concat.  + ATV}'' indicates the early fusion of the three visual features that are then fed into the trainable ATV sub-network of \cite{galanopoulos2020} (as illustrated in the rightmost part of Fig. \ref{fig_1}a), followed by the required FC layers in order to encode the ATV's output into the corresponding joint  feature spaces. Similarly, ``\textit{feat. concat.}'' refers to the early fusion of the three visual features (as illustrated in the rightmost part of Fig. \ref{fig_1}b) followed by the required FC layers. Finally, ``\textit{only FCs}'' refers to the proposed strategy (Fig. \ref{fig_1}c), where visual features are individually and directly encoded to the joint spaces using only FC layers.

\begin{table}[]
\caption{Comparison of combinations of textual encoders and visual features on the IACC.3 and V3C1 datasets}
\label{tab:tabavs}
\resizebox{\columnwidth}{!}{%
\begin{tabular}{lc|c|lll|lll|l}
\hline
\multicolumn{2}{c}{Model} &\multicolumn{1}{c}{\multirow{4}{*}{\begin{tabular}[c]{@{}c@{}} \# of feature \\ spaces\end{tabular}}}  & \multicolumn{6}{c}{Datasets (all scores: MxinfAP)} &  \\ \cline{1-9}

\multicolumn{1}{c|}{\multirow{2}{*}{\begin{tabular}[c]{@{}c@{}}Visual features \\ combination strategy\end{tabular} }} & \multicolumn{1}{c|}{\multirow{2}{*}{Textual encoders (Fig. \ref{fig_1}c)}} &\multicolumn{1}{c|}{}  & \multicolumn{3}{c|}{IACC.3} & \multicolumn{3}{c}{V3C1} &  \\ \cline{4-10} 

\multicolumn{1}{c|}{} & \multicolumn{1}{c|}{} & \multicolumn{1}{c|}{} & \multicolumn{1}{c}{AVS16} & \multicolumn{1}{c}{AVS17} & \multicolumn{1}{c|}{AVS18} & \multicolumn{1}{c}{AVS19} & \multicolumn{1}{c}{AVS20} & \multicolumn{1}{c|}{AVS21} & \multicolumn{1}{c}{Mean} \\ \hline

\multicolumn{1}{c|}{\multirow{4}{*}{ \begin{tabular}[c]{@{}c@{}}(feat. concat.  + ATV)\\ $\downarrow$ \\FCs \\(Fig. \ref{fig_1}a)\end{tabular} }} & bow, w2v, bert, bi-gru &   4 & 0.169  & 0.250 & 0.126  &   0.170 & 0.252	  & 0.251 	  &  0.203 \\
\multicolumn{1}{l|}{} &  bow, w2v, bert, bi-gru, clip &  5 &  0.179 & 0.265 	  & 0.133  &  0.174  & 0.260	  & 0.254 	  & 0.211  \\
\multicolumn{1}{l|}{} & w2v-bert, clip, ATT & 3 & 0.199	 & 0.265 &  0.135 & 0.183  &  	0.271 &  0.262	 &  0.219\\
\multicolumn{1}{l|}{} &  clip, ATT & 2 & 0.211 &  	0.288 & 	0.143  &  0.188 & 	0.288  & 	0.271	  & 0.232  \\
\hline

\multicolumn{1}{c|}{\multirow{4}{*}{ \begin{tabular}[c]{@{}c@{}}(feat. concat.) \\ $\downarrow$ \\FCs \\(Fig. \ref{fig_1}b)\end{tabular} }} &  bow, w2v, bert, bi-gru &  4 & 0.184  & 0.267   & 0.136   & 0.202   & 0.303	  & 0.295 	  &  0.231 \\
\multicolumn{1}{l|}{} &  bow, w2v, bert, bi-gru, clip &  5  &  0.198 & 0.275 	  & 0.141  & 0.205   & 0.305	  & 0.292  	  & 0.236  \\
\multicolumn{1}{c|}{} &  w2v-bert, clip, ATT & 3 & 0.210 &  	0.281	 & 0.147  &  0.194 & 	0.290 & 	0.275	 & 0.233 \\
\multicolumn{1}{l|}{} & clip, ATT & 2 & 0.227	 &  0.291 & 	0.149 &  0.190 & 	0.296	  &  0.292	 & 0.241 \\ 
\hline

\multicolumn{1}{c|}{\multirow{4}{*}{\begin{tabular}[c]{@{}c@{}}Only FCs \\(Fig. \ref{fig_1}c) \end{tabular}}} &  bow, w2v, bert, bi-gru & 12  & 0.191  &  0.294 & 0.150  & 0.202   & 0.303	  & 0.295 	  & 0.239  \\
\multicolumn{1}{l|}{} &  bow, w2v, bert, bi-gru, clip &  15 &  0.205 & 0.306 	  & 0.152  &  0.205  & 0.305	  & 0.292 	  &  0.244 \\
\multicolumn{1}{c|}{} & w2v-bert, clip, ATT  & 9 & 0.219		 &  0.312	  & 0.150   & 0.210 & 0.307  & 	0.297 & 	0.249 \\
\multicolumn{1}{l|}{} & clip, ATT & 6 &  \textbf{0.234}	 & \textbf{0.317} &	\textbf{0.153}  &  \textbf{0.220} & 	\textbf{0.316}  & 	\textbf{0.312}& 	\textbf{0.259 }\\
\hline

\end{tabular}%
}
\end{table}

\begin{table}[]
\caption{Comparison of combinations of textual encoders and visual features on the full and 1k-A variations of the MSR-VTT dataset}
\label{tab:tabmsr_2}
\resizebox{\columnwidth}{!}{%

\begin{tabular}{cc|cccccccccc}
\hline
\multicolumn{2}{c}{\multirow{2}{*}{Model}}               & \multicolumn{10}{c}{Datasets}                                                                   \\ \cline{3-12} 
\multicolumn{2}{c}{}                                     & \multicolumn{5}{c|}{MSR-VTT full}                                   & \multicolumn{5}{c}{MSR-VTT 1k-A}            \\ \hline
\multicolumn{1}{c|}{\begin{tabular}[c]{@{}c@{}}Visual features \\ combination strategy\end{tabular}}   & Textual encoders  (Fig. \ref{fig_1}c)& R@1$\uparrow$ & R@5$\uparrow$ & R@10$\uparrow$ & Medr$\downarrow$ & \multicolumn{1}{c|}{mAP$\uparrow$} &  R@1$\uparrow$ & R@5$\uparrow$ & R@10$\uparrow$ & Medr$\downarrow$ & mAP$\uparrow$  \\ \hline

\multicolumn{1}{c|}{\multirow{4}{*}{\begin{tabular}[c]{@{}c@{}}(feat. concat.  + ATV)\\ $\downarrow$ \\FCs \\(Fig. \ref{fig_1}a)\end{tabular}}} &  bow, w2v, bert, bi-gru & 17.3    & 40.9    &    52.9  & 9     & \multicolumn{1}{c|}{28.9}         & 32.2    & 62.9    &    73.7  & 3     & 46.3     \\

\multicolumn{1}{c|}{} &  bow, w2v, bert, bi-gru, clip    &  19.2   &    43.6 &  55.6    & 8     & \multicolumn{1}{c|}{30.1}     &  34.8   &   63.9  &  75.8    &  3    &  48.6     \\
\multicolumn{1}{l|}{} & w2v-bert, clip, ATT   &   21.4  & 46.6    & 58.5     & 7     &  \multicolumn{1}{c|}{33.4}     &  36.5 &  67.4    &  76.4      &     2 &     50.3    \\
\multicolumn{1}{l|}{} &  clip, ATT   &  \textbf{22.3}   &    \textbf{47.6}&\textbf{59.3}    & \textbf{6}     &  \multicolumn{1}{c|}{\textbf{34.3}}     & \textbf{40.1} &   \textbf{68.4}&   \textbf{78.1}    & \textbf{2}     &  \textbf{52.9}     \\
\hline

\multicolumn{1}{c|}{\multirow{4}{*}{ \begin{tabular}[c]{@{}c@{}}(feat. concat.) \\ $\downarrow$ \\FCs \\(Fig. \ref{fig_1}b)\end{tabular} }} &  bow, w2v, bert, bi-gru  & 17.6    & 41.3    &  53.2    &  9    &  \multicolumn{1}{c|}{29.0}     &   32.8  & 64.4    &    74.3  & 3     & 47.1        \\
\multicolumn{1}{l|}{} &  bow, w2v, bert, bi-gru, clip    &  19.7   & 44.1     &     55.9 & 8     & \multicolumn{1}{c|}{31.4}     &  35.4   &  65.0   & 76.0     &  3    & 48.5    \\
\multicolumn{1}{l|}{} & w2v-bert, clip, ATT   &  21.4   & 46.3    & 57.7     & 7     &  \multicolumn{1}{c|}{33.3}     &  35.7   & 67.4    &    77.4  & 3     & 49.8         \\
\multicolumn{1}{l|}{} &  clip, ATT   &  22.0   &    46.5 & 57.9     & 7     & \multicolumn{1}{c|}{33.7}     &  37.1   &    65.4 &  75.9    & 2     & 50.3     \\
\hline

\multicolumn{1}{c|}{\multirow{4}{*}{\begin{tabular}[c]{@{}c@{}}Only FCs \\(Fig. \ref{fig_1}c) \end{tabular}}}  & bow, w2v, bert, bi-gru  &  17.1   &     40.9 & 53.0     & 9     &   \multicolumn{1}{c|}{28.6}     & 33.0    & 63.2    & 74.1     & 3     & 46.9         \\
\multicolumn{1}{l|}{} &  bow, w2v, bert, bi-gru, clip    &  18.9   &  43.6   &  55.6    &  8    &   \multicolumn{1}{c|}{30.8}     &      35.2   &  65.2    & 76.3     &    3 &   48.9   \\
\multicolumn{1}{l|}{} & w2v-bert, clip, ATT   &  20.9   &   45.6  &  57.7    &   7   &  \multicolumn{1}{c|}{32.6}     &   35.8  & 66.3    &77.2      & 3     & 49.6         \\
\multicolumn{1}{l|}{} &  clip, ATT   & 21.2    &  46.3   &  58.2    &   7   &   \multicolumn{1}{c|}{33.1}     &  36.5   &   66.9  &  77.7    &   2   &  50.2  \\
\hline
 
\end{tabular}
}
\end{table}

In Table \ref{tab:tabavs} we report the results on the AVS datasets when using different combinations of textual encoders together with the aforementioned three possible visual encoding strategies. Concerning the visual encoding strategies, the results indicate that the lowest performance is achieved by the ``\textit{feat. concat.  + ATV}'' strategy, regardless of the textual encoders choice. When the early fusion of the trained models ``\textit{feat. concat.}'' is used instead of ATV, the performance consistently increases. The best results are achieved by forwarding the visual features independently with FC layers, as in the proposed approach. 
Regarding the combinations of the textual encoders, we can see that the utilization of fewer but more powerful encoders (i.e. clip and ATT) leads to better results than using a multitude of, possibly weak, encoders as in \cite{li2020sea}, regardless of the employed visual encoding strategy.  These evaluations show that {\it how} the textual and visual features are combined has significant impact on the obtained results.

 In Table \ref{tab:tabmsr_2} we presented the same ablation study for the full and 1k-A variations of the MSR-VTT dataset. In these datasets, we can observe a different behavior concerning the visual encoding: while the utilization of a few and powerful textual encoders (i.e. clip and ATT) continues to perform the best, when it comes to the visual modality, the ``\textit{feat. concat.  + ATV}'' strategy consistently performs the best, regardless of the textual encoder choice. This finding, combined with the results of Table \ref{tab:tabavs}, shows that for similar yet different problems and datasets there is no universally-optimal way of combining the visual features (and one can reasonably assume that this may also hold for the textual ones).

\section{Conclusions}
In this work, we presented a new network architecture for efficient text-to-video retrieval. We experimentally examined different combinations of visual and textual features, and concluded that selectively combining the textual features into fewer but more powerful textual encoders leads to improved results. Moreover, we shown how a fixed set of background queries extracted from large-scale captioning datasets can be used together with softmax operations at the inference stage for revising query-video similarities, leading to improved video retrieval. Extensive experiments and comparisons on different datasets document the value of our approach.

\paragraph{\textbf{Acknowledgements.}} This work was supported by the EU Horizon 2020 programme under grant agreements H2020-101021866 CRiTERIA and H2020-832921 MIRROR.


%
%

\end{document}